\documentclass[letterpaper]{article} % DO NOT CHANGE THIS
\usepackage{aaai24}  % DO NOT CHANGE THIS
\usepackage{times}  % DO NOT CHANGE THIS
\usepackage{helvet}  % DO NOT CHANGE THIS
\usepackage{courier}  % DO NOT CHANGE THIS
\usepackage[hyphens]{url}  % DO NOT CHANGE THIS
\usepackage{graphicx} % DO NOT CHANGE THIS
\usepackage{natbib}  % DO NOT CHANGE THIS AND DO NOT ADD ANY OPTIONS TO IT
\usepackage{caption} % DO NOT CHANGE THIS AND DO NOT ADD ANY OPTIONS TO IT
\usepackage{algorithm}
\usepackage{algorithmic}

\usepackage{newfloat}
\usepackage{listings}
\DeclareCaptionStyle{ruled}{labelfont=normalfont,labelsep=colon,strut=off} % DO NOT CHANGE THIS
\floatstyle{ruled}
\newfloat{listing}{tb}{lst}{}
\floatname{listing}{Listing}
\title{KAM-CoT: Knowledge Augmented Multimodal Chain-of-Thoughts Reasoning}
\author{
    Debjyoti Mondal, Suraj Modi, Subhadarshi Panda,\\
    Rituraj Singh, Godawari Sudhakar Rao
}
\affiliations{
    Samsung R\&D Institute India - Bangalore\\
    \{d.mondal, suraj.modi, subha.darshi, rituraj.s, g.sudhakar\}@samsung.com
}

\usepackage{bibentry}
\usepackage{tablefootnote}
\usepackage{amsfonts}
\usepackage{amsmath}
\usepackage{amssymb}
\usepackage{pifont}

\begin{document}

\maketitle

\begin{abstract}
Large Language Models (LLMs) have demonstrated impressive performance in natural language processing tasks by leveraging chain of thought (CoT) that enables step-by-step thinking. Extending LLMs with multimodal capabilities is the recent interest, but incurs computational cost and requires substantial hardware resources. To address these challenges, we propose KAM-CoT a framework that integrates CoT reasoning, Knowledge Graphs (KGs), and multiple modalities for a comprehensive understanding of multimodal tasks. KAM-CoT adopts a two-stage training process with KG grounding to generate effective rationales and answers. By incorporating external knowledge from KGs during reasoning, the model gains a deeper contextual understanding reducing hallucinations and enhancing the quality of answers. This knowledge-augmented CoT reasoning empowers the model to handle questions requiring external context, providing more informed answers. Experimental findings show KAM-CoT outperforms the state-of-the-art methods. On the ScienceQA dataset, we achieve an average accuracy of 93.87\%, surpassing GPT-3.5 (75.17\%) by 18\% and GPT-4 (83.99\%) by 10\%. Remarkably, KAM-CoT achieves these results with only 280M trainable parameters at a time, demonstrating its cost-efficiency and effectiveness.
\end{abstract}

\section{Introduction}
Large Language Models (LLMs), particularly GPT-3 \citep{kojima2022large}, 
% PaLM \citep{chowdhery2022palm}, 
ChatGPT \citep{chatgpt} and recently LLaMA, LLaMA2 \citep{touvron2023llama1,touvron2023llama} 
have demonstrated exceptional performance in natural language processing tasks. Additionally, 
incorporation of chain of thought (CoT) method in LLMs has revolutionized 
the way machines approach reasoning intensive tasks \citep{zhou2022least}. 
CoT refers to the ability of LLMs to think and reason in a step-by-step manner, 
mirroring the human cognitive processes \citep{wei2022chain}. Traditional language models (LMs) generate
responses without explicit intermediate steps, which may lead to sub-optimal answers, especially 
in complex reasoning scenarios. CoT addresses the limitations by enabling language models to 
reason by introducing intermediate steps, thereby enhancing the model's problem-solving capabilities. 

\begin{figure}[]
    \begin{center}
        \includegraphics[width = 239pt]{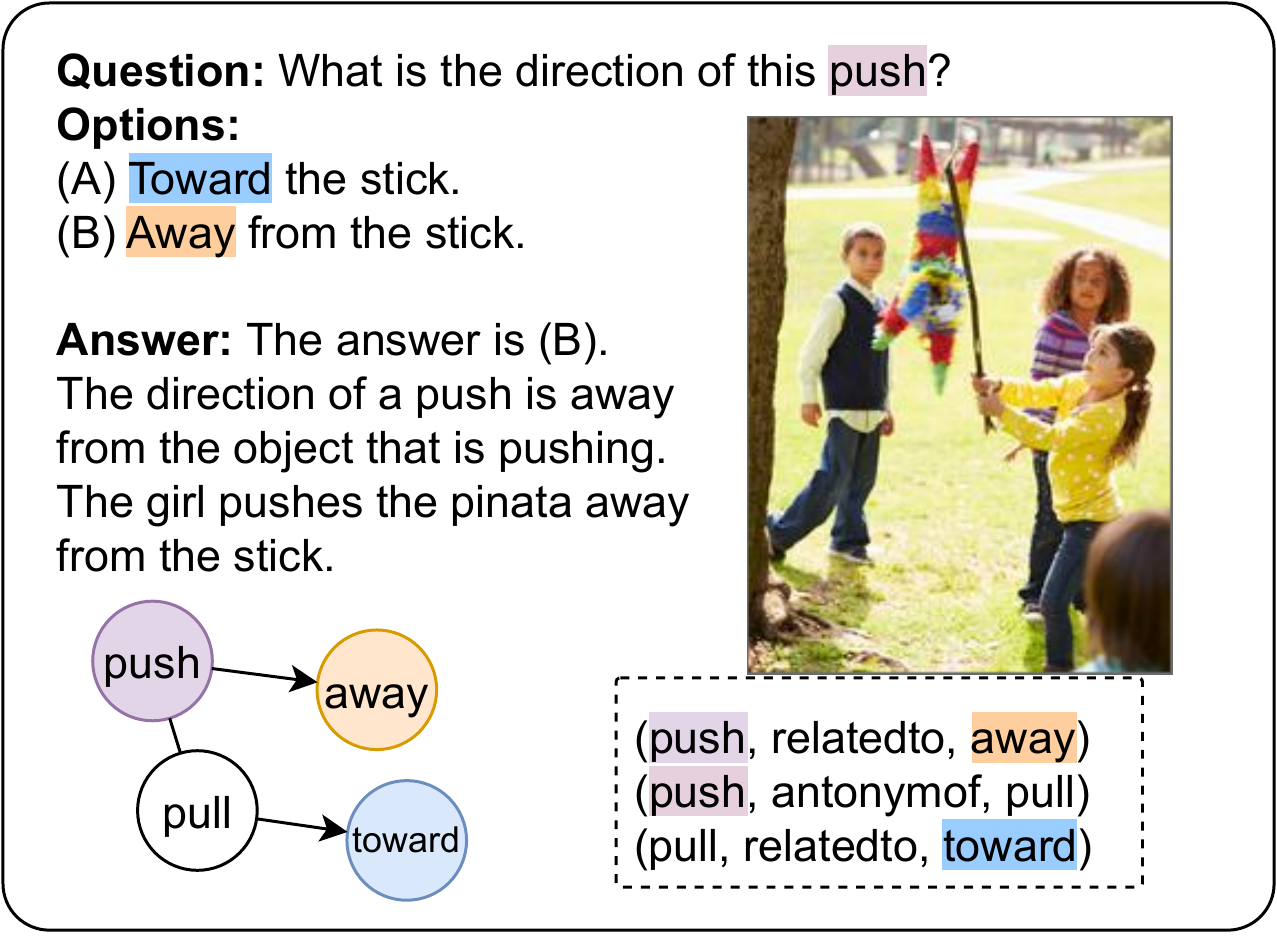}
    \end{center}
    \caption{An example from ScienceQA dataset~\cite{lu2022learn} showing how graphs can aid in multi-modal QA.}
    \label{figure:toy}
\end{figure}

Recently, there is a surge to extend LLMs with multimodal capabilities. 
% Multimodal learning resides at the intersection of computer vision and natural language processing. 
The fusion of visual and textual 
information has led to significant advancements in vision-and-language tasks, like visual question answering (VQA), image captioning, and image-text retrieval, and has opened up potential for transformative progress.
Authors \citet{liu2023visual, gao2023llama, lu2023chameleon} recognize and 
advocate the value of amalgamating visual and
linguistic modalities. However, the behemoth scale of these models necessitates 
substantial computational resources, particularly in terms of hardware 
infrastructure. \citet{zhang2023multimodal} proposes fine-tuning smaller models 
to adapt to multimodality and elicit CoT capabilities. Nevertheless, such an approach tends to result in 
hallucinations, where the model generates plausible, but incorrect reasoning 
and answers. One possible solution is to integrate Knowledge Graphs (KGs) for enhancing 
model comprehension.

KGs serve as valuable structured knowledge sources, 
capturing information from various domains. For CoT reasoning, KGs 
can supplement step-by-step reasoning. By incorporating 
information from KGs, language models can reason more coherently, 
and leverage contextual relationships between entities and attributes. Consider the question in Figure~\ref{figure:toy}. The knowledge about the direction of push is pivotal to answer the question. The KG triples (shown in bottom-right corner in Figure~\ref{figure:toy}) about object relationship and orientation, equips the model to answer correctly. The integration 
enhances the quality of generated responses, especially in tasks that require
complex reasoning and context-aware understanding.

In this work, we propose to augment multiple modalities with knowledge graphs to help the
model solve complex problems eliciting CoT capabilities. The proposed approach, KAM-CoT, consists of an LM that takes language context, a vision encoder to encode visual features and a graph
neural network (GNN) that reasons over the KGs. Following \citet{zhang2023multimodal}, we decouple the reasoning process into two sequential stages. 
In the first stage, we generate well-reasoned rationales. 
The second stage takes the generated rationale as an additional input and provides answers. 
KAM-CoT seamlessly stitches text, vision and graph features together, enabling machines to think 
and reason coherently, similar to human cognition. We evaluate our proposed model 
on the ScienceQA \citep{lu2022learn} benchmark. We achieve an average accuracy of 93.87\%, 
surpassing GPT-3.5 (75.17\%) by 18\% and GPT-4 (83.99\%) by 10\%. Additionally, KAM-CoT 
achieves these results with only 280M trainable parameters at a time, demonstrating its 
cost-efficiency and effectiveness.

This paper makes the following contributions:\\
\textbf{1. Graph Extraction:} We extract salient triples from ConceptNet \citep{speer2017conceptnet} based on the given question context.\\
\textbf{2. Fusion with KG:} We propose a few indicative mechanisms for fusing text and image modalities with the knowledge graph, and examine their efficiency.\\
\textbf{3. KAM-CoT:} We propose the Knowledge Augmented Multimodal CoT approach, KAM-CoT. The 280M model jointly processes vision, text, and knowledge graph in stages, does step-by-step reasoning to generate plausible reasoning and answers.

We conduct extensive experiments and evaluation on the ScienceQA dataset\cite{lu2022learn}, 
achieving new state-of-the-art performance. 
We also look into the effects and contributions of each component and
discuss potential directions for future research.

\section{Related Work}
We explore related works in four key areas: in-context learning, 
CoT through fine-tuning approaches, vision-language models and knowledge augmented methods. 

\paragraph{In-context learning} 
%LLMs \citep{zhao2023survey} have exhibited remarkable abilities in generating contextually coherent text. 
LLMs~\citep{zhao2023survey} exhibit the capability of CoT through two principal modes: Zero shot and Few shot. 
Zero shot performs inference 
without necessitating any explicit examples or guidance. 
Recent studies have revealed that LLMs can achieve satisfactory results when 
prompted with the phrase ``Let's think step by step" \citep{kojima2022large}. In few 
shot context, LLMs are provided with a set of demonstrative examples that serve 
as guides, enabling them to grasp and learn patterns from these instances. The examples are 
curated by human experts.

Auto-CoT introduces the automatic construction of demonstration examples 
using LLMs \citep{zhang2023automatic}. It generates examples with inherent noise. With automatic sampling 
of diverse questions and post-processing quality control mechanisms, it gets usable chains. 
% Self-Taught Reasoner (STaR) \citep{zelikman2022star} proposes 
% a loop based system to generate correct rationales. 
\citet{wang2022self} proposes a 
decoding self-consistent strategy that samples from a 
diverse set of reasoning paths and subsequently selects the most consistent answer by 
marginalizing all possible paths.
PROMPTPG~\citep{lu2022dynamic} employs policy gradient techniques to acquire the ability 
to discern contextually related examples from the limited set of training samples and then 
construct the corresponding prompt for a given sample. \citet{chen2022program} proposes 
\textit{Program of Thoughts}, where the computation is delegated to an interpreter, 
decoupling complex computation from reasoning and understanding. Another interesting work, \textit{least-to-most prompting} \citep{zhou2022least} proposes to break a complex problem into simpler ones and solve them sequentially by leveraging 
the answer from previously solved sub-problems. However, all these approaches are limited to 
LLMs, reasonably greater than 100B parameters~\citep{wei2022emergent}.

\paragraph{CoT through fine-tuning approaches} \citet{lu2022learn} proposes a Science Question-Answer (ScienceQA) dataset that consists of multimodal multiple choice questions 
with corresponding lectures, explanations and correct answers. Authors observe
improvements in question answering by using CoT by 1.20\% in few shot GPT-3 and 
3.99\% in fine-tuned UnifiedQA \citep{khashabi-etal-2020-unifiedqa}. MM-CoT \citep{zhang2023multimodal} 
proposes to fine-tune an LM on ScienceQA dataset with CoT method. They propose rationale 
generation and answer inference in two stages. The model 
outperforms GPT-3.5 by 16\% on this dataset and surpasses human performance. 

\paragraph{Vision-Language Models} With the proposal of visual question 
answering tasks \citep{antol2015vqa}, there have been plenty of works in 
aligning vision and language modalities. ViLT \citep{kim2021vilt} proposes a single 
transformer architecture for text and image modalities that facilitates seamless 
cross modal interaction. Patch-TRM (Transformer with cross-modal TRM) parses 
images into ordered patches in a hierarchical pyramid layout \citep{lu2021iconqa}. 
The patches are encoded with pre-trained ResNet and passed through a vision transformer. 
VisualBERT proposes a unified architecture that leverages the expressive 
power of transformer based BERT model and aligns the features extracted from 
images \citep{li2019visualbert, li-etal-2020-bert-vision}. In particular, both visual 
and textual inputs are masked, and the model learns to predict the masked inputs, 
enabling it to capture contextual alignment. 
BLIP2 \citep{li2023blip} proposes QFormer, pretrained with a
two-stage strategy to align image encoders and LLMs.
\citet{liu2023prismer} 
proposes the Prismer model, that uses an ensemble of domain experts. 
KOSMOS \citep{huang2023language} trains a model from scratch on web-scale multimodal
corpora, including arbitrarily interleaved text and images, image-caption pairs, and text data.

Recently with the advent of LLaMA models, there has been significant progress 
in instruction-following language modelling. LLaVA \citep{liu2023visual} relies 
on the text-only GPT-4 \citep{openai2023gpt4} model, to generate multimodal 
data. The authors propose two stage training: pre-training for 
feature alignment and instruction-following fine-tuning. LLaMA-Adapter V2 \citep{gao2023llama} 
proposes a parameter-efficient adapter based visual instruction model that distributes instruction 
following ability across the entire model. LaVIN \citep{luo2023cheap} is
another parameter-effecient technique based on mixture 
of modalities. SCITUNE \citep{horawalavithana2023scitune} and T-SciQ \citep{wang2023tsciq} 
are science-focused visual and language understanding models. Chameleon \citep{lu2023chameleon} 
mitigates the limitations of accessing up-to-date information, by augmenting LLMs with 
plug-and-play modules for compositional reasoning. However all these instruction 
following methods require larger models, usually greater than
7B parameters.

\paragraph{Knowledge augmented methods} Several recent 
studies have explored infusion of structured knowledge into LMs.
SKILL \citep{moiseev-etal-2022-skill} proposes conversion of
KG triples into sentences and then using them for pretraining. 
KagNet \citep{lin2019kagnet} proposes to ground a question-answer pair from the 
semantic space to the knowledge-based symbolic space as a schema graph, and then 
trains a graph convolution network with a hierarchical path-based attention mechanism. 
QA-GNN \citep{yasunaga2021qa} proposes the use of LMs to estimate the importance of 
nodes in a KG with respect to the given context, and does joint reasoning over 
a unified graph. \citet{Zhang2022GreaseLMGR} proposes the GreaseLM model that fuses encoded 
representations from pretrained LMs and graph neural networks over multiple layers 
of language-KG interaction. Extending to multiple modalites, VQA-GNN \citep{wang2022vqa} 
proposes to unify the image-level scene graph with conceptual knowledge to perform 
joint reasoning over the unified graph.

\section{Method}
\label{sec:method}

\begin{figure}[t!]
    \centering
    \includegraphics[width = 239pt]{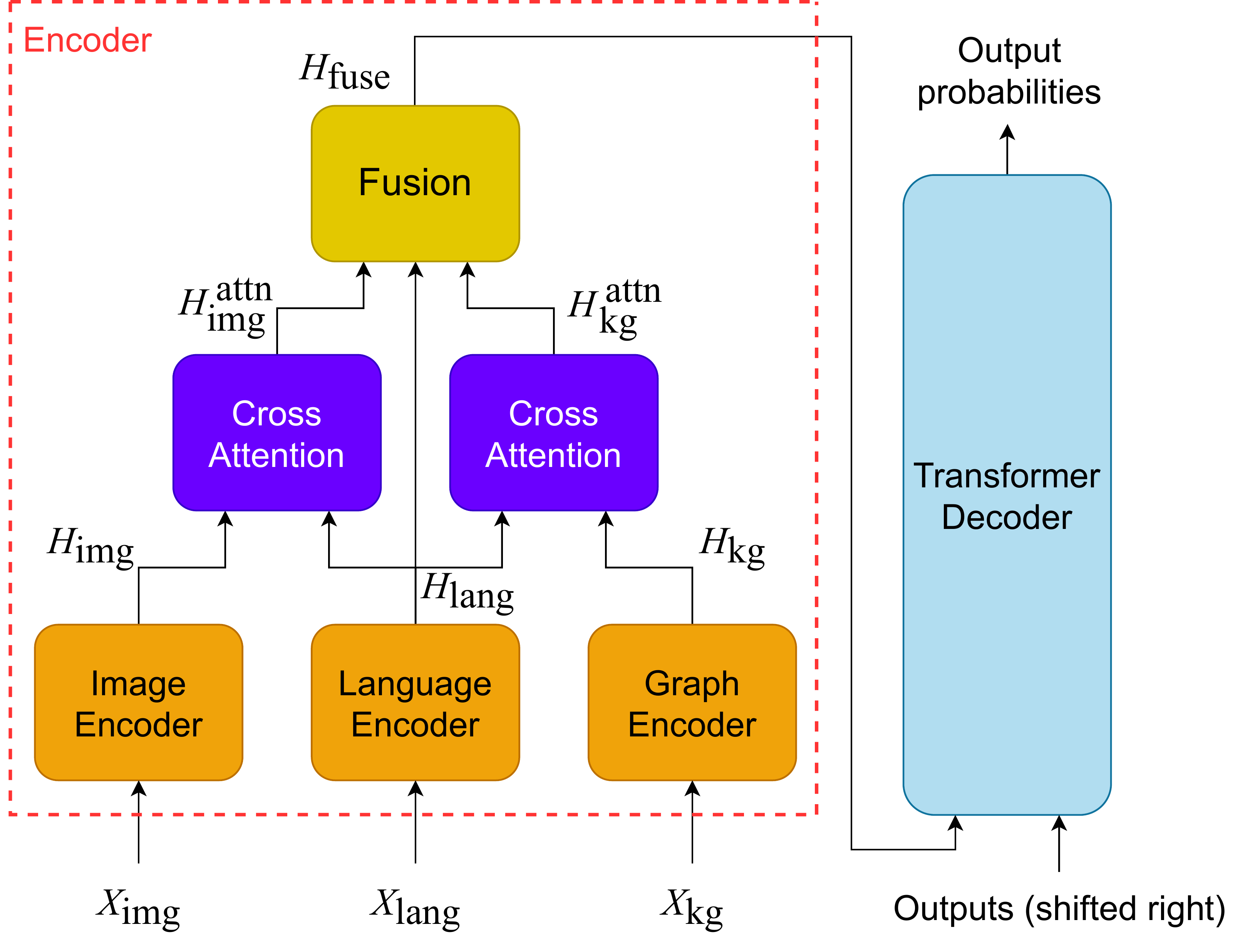}
    \caption{KAM-CoT model architecture.}
    \label{fig:model_architecture}
\end{figure}

We describe the proposed KAM-CoT approach in this section.
As an overview, KAM-CoT involves encoding the 
language, image and the graph input.
Note that the graph is derived from the language input.
The three modalities are then made to interact with each other using cross-attention.
Finally, the fused features are fed to a transformer decoder that 
generates text autoregressively.

\subsection{Task Formulation}

Given a question $q$ along with $k$ answer 
choices $\{a_1, a_2, \hdots, a_k\}$, the task is to pick the correct choice.
The question $q$ is optionally accompanied by an 
image $X_{\text{img}}$ and a text $c$ that adds context to it.

One potential approach is to use a neural network to 
generate the right choice directly.
However, as already established, chain-of-thoughts reasoning helps in 
inferring the right answer, especially for complex reasoning 
tasks \citep{wei2022chain, NEURIPS2022_8bb0d291}.
We therefore train the model to generate a rationale $r$ for 
the answer, in the first step.
The next step involves picking the correct answer by 
conditioning the generation process on $r$,
along with the existing inputs.
The rationale generation and answer identification models 
are the same,
but they are trained separately from identical initializations.
This is similar to the technique used by \citet{zhang2023multimodal} 
who deal with just image and text modalities.
In our case, we extend their approach to handle 
graphs as an additional modality that would ground 
the generation process on factual knowledge.

To obtain the language input for rationale 
generation, we simply concatenate the different text 
portions, $X_{\text{lang}}^{\text{rat}} = [q; c; [a_1, a_2, \hdots, a_k]]$.
And for answer choice prediction, 
we append the rationale $r$ as well to 
obtain $X_{\text{lang}}^{\text{ans}} = [q; c; [a_1, a_2, \hdots, a_k]; r]$.

We extract a subgraph $X_{\text{kg}}$ for each sample (discussed in details below).
For rationale generation, we learn a model $F_{\text{rat}}(.)$ that generates the rationale $r$.
\begin{align}
    r = F_{\text{rat}}(X_{\text{lang}}^{\text{rat}}, X_{\text{img}}, X_{\text{kg}})
\end{align}
Similarly, for generating text to identify the right answer, we learn a model $f_{\text{ans}}(.)$.
\begin{align}
    a = F_{\text{ans}}(X_{\text{lang}}^{\text{ans}}, X_{\text{img}}, X_{\text{kg}})
\end{align}

Formalizing the procedure, with the modalities given 
to the model as input, we compute and maximize the probability 
of generating the reference text $Y$, which can either 
be the rationale or the answer, of length $N$.
\begin{align*}
    p(Y | X_{\text{lang}}, X_{\text{img}}, X_{\text{kg}}) = \prod_{i=1}^{N}p_{\theta}(Y_i | X_{\text{lang}}, X_{\text{img}}, X_{\text{kg}}, Y_{<i})
\end{align*}
The model $p_{\theta}$ is made with a combination of a graph encoder and a transformer network.
Algorithm \ref{algo:kacot} lists the steps involved in the KAM-CoT algorithm.

\subsection{Encode Inputs From Different Modalities}

\begin{algorithm}[t!]
    \caption{KAM-CoT Reasoning}
    \label{algo:kacot}
    \textbf{Input:} Language features $X_{\text{lang}}^{\text{rat}}$, Image features $X_{\text{img}}$, and Graph features $X_{\text{kg}}$ \\
    \textbf{Output:} Rationale $r$, Answer $a$
    \begin{algorithmic}[1]
        \STATE Construct input $X = \{X_{\text{lang}}^{\text{rat}}, X_{\text{img}}, X_{\text{kg}}\}$
        \STATE $r \gets F_{\text{rat}}(X)$
        \STATE Concatenate $r$ to $X_{\text{lang}}^{\text{rat}}$, to make $X_{\text{lang}}^{\text{ans}} \gets [X_{\text{lang}}^{\text{rat}}; r]$
        \STATE Construct new input $X' = \{X_{\text{lang}}^{\text{ans}}, X_{\text{img}}, X_{\text{kg}}\}$
        \STATE $a \gets F_{\text{ans}}(X')$
        % \item[]
        \STATE \textbf{procedure} $F(X)$
        \STATE \quad Get the encoded representations, $H_{\text{lang}}$, $H_{\text{img}}$, and $H_{\text{kg}}$
        \STATE \quad Obtain the feature representations, $H_{\text{img}}^{\text{attn}}$, and $H_{\text{kg}}^{\text{attn}}$
        \STATE \quad Fuse these representations with $H_{\text{lang}}$ to obtain $H_{\text{fuse}}$
        \STATE \quad Input $H_{\text{fuse}}$ to the decoder to get the target $Y$
        \STATE \quad \textbf{return} $Y$
        \STATE \textbf{end procedure}
    \end{algorithmic}
\end{algorithm}

\subsubsection{Text Encoding}

We use a transformer based language encoder to encode 
$X_{\text{lang}}$ to obtain $H_{\text{lang}} = \text{LanguageEncoder}(X_{\text{lang}}) \in \mathbb{R}^{n \times d}$, 
where $n$ is the number of tokens in $X_{\text{lang}}$ and $d$ is 
the output embedding size of the language encoder.

\subsubsection{Image Encoding}

We encode the image $X_{\text{img}}$ using a transformer based 
image encoder to obtain $H_{\text{img}} = \text{ImageEncoder}(X_{\text{img}})W_{\text{img}} \in \mathbb{R}^{m \times d}$ where $m$ is the number of patches in the image.
The projection matrix $W_{\text{img}}$ brings the output 
embedding dimension to $d$, same as that of $H_{\text{lang}}$. 

\subsubsection{Subgraph Extraction}

For every sample, we extract a subgraph from
ConceptNet \citep{speer2017conceptnet} by following a method 
similar to that in \citet{yasunaga2021qa}.
% ConceptNet consists of head, relation, and tail triples.
We group the relations in ConceptNet into $17$ distinct types. 
These relations can be either forward or backward, yielding a total of $34$ possible edge types.
The triples are converted to sentences, and corresponding sentence patterns are stored.
These patterns are used to ground and extract nodes from the question, 
context and answer choices.
A subgraph is made of (i) $\mathbf{V}$, a set of nodes, (ii) $\mathbf{E}$, a set of edges 
and (iii) $\boldsymbol{\mathbf{\phi}}$, a function which maps every edge to an integer in the range 
$[0, 33]$, representing the edge type.
To get the initial node embeddings, 
we the same pretrained checkpoint of the language encoder 
used for text encoding, and average the embeddings over the 
span of all occurences of that node \citep{feng-etal-2020-scalable}.
The thought behind using the same language encoder checkpoint 
is to ensure that the language and node embeddings start from the same space.\footnote{We also experiment with using image captions for grounding. In that case, we simply append the caption to the existing context.}
Let $N_{\text{qa}}$ represent this set of grounded nodes.
For every pair of nodes, $n_{\text{a}}, n_{\text{b}} \in N_{\text{qa}}$, we 
append all common nodes in their 1-hop neighbourhood into $N_{\text{1-hop}}$.
We repeat this process for each pair of nodes 
in $N_{\text{qa}}$ and $N_{\text{1-hop}}$ and append 
the nodes into $N_{\text{2-hop}}$.
This way, we get a graph connecting all nodes 
in $N_{\text{qa}}$ to each other with a path length of 
atmost 2 intermediate nodes: $\mathbf{V} = N_{\text{qa}} \cup N_{\text{1-hop}} \cup N_{\text{2-hop}}$.
Since the number of nodes could grow exponentially, we follow the 
pruning strategy in \citet{yasunaga2021qa} to keep the 
top 200 nodes for every sample.
For the edges, we build an embedding table and learn embeddings during training.

\subsubsection{Graph Encoding}

Using a combination of graph layers, we encode the extracted subgraph 
$X_{\text{kg}}$ to obtain the node 
embeddings $H_{\text{kg}} = \text{KGEncoder}(X_{\text{kg}}) \in \mathbb{R}^{p \times d}$, 
where $p$ is the number of extracted nodes. 

\subsection{Interaction Between Modalities}

We use cross-attention to enable the interaction between the representations of text, image and subgraph.
For this we use two seperate single-headed attention modules (see Figure \ref{fig:model_architecture}). 
For the first attention module, the language and image embeddings interact. 
Similarly, in another attention module interaction between language and node 
embeddings happen.
\begin{align}
    H_{\text{img}}^{\text{attn}} = \text{softmax}\Big(\frac{H_{\text{lang}}H_{\text{img}}^\top}{\sqrt{d}}\Big)H_{\text{img}}
\end{align}
\begin{equation}
    H_{\text{kg}}^{\text{attn}} = \text{softmax}\Big(\frac{H_{\text{lang}}H_{\text{kg}}^\top}{\sqrt{d}}\Big)H_{\text{kg}}
\end{equation}
% We therefore obtain the `attended' image and node embeddings.

\subsection{Fusion}
\label{sec:fusion}

We use gated fusion \citep{wu-etal-2021-good,Zhang2020NeuralMT,li-etal-2022-vision} to get the final representation.
\begin{equation}
    \begin{split}\label{eq:fuse1}
        S_{\alpha} = H_{\text{lang}}W_{1} + H_{\text{img}}^{\text{attn}}W_{2} + H_{\text{kg}}^{\text{attn}}W_{3} \in \mathbb{R}^{n \times d}
        \\
        S_{\beta} = H_{\text{lang}}W_{4} + H_{\text{img}}^{\text{attn}}W_{5} + H_{\text{kg}}^{\text{attn}}W_{6} \in \mathbb{R}^{n \times d}
        \\
        S_{\gamma} = H_{\text{lang}}W_{7} + H_{\text{img}}^{\text{attn}}W_{8} + H_{\text{kg}}^{\text{attn}}W_{9} \in \mathbb{R}^{n \times d}
    \end{split}\end{equation}
    \begin{equation}\begin{split}\label{eq:fuse2}
        \alpha_{ij}, \beta_{ij}, \gamma_{ij} = \text{softmax}([S_{\alpha_{ij}}, S_{\beta_{ij}}, S_{\gamma_{ij}}])
        \\    
        H_{\text{fuse}} = \alpha \cdot H_{\text{lang}} + \beta \cdot H_{\text{img}}^{\text{attn}} + \gamma \cdot H_{\text{kg}}^{\text{attn}} \in \mathbb{R}^{n \times d}
    \end{split}
\end{equation}
Here $\alpha, \beta, \gamma \in [0, 1]^{n \times d}$ and sum to 1 element-wise, 
and all $W \in \mathbb{R}^{d \times d}$. We will refer to this fusion method as \textbf{Fusion-1}.
We discuss and compare a few other fusion variants in the Discussion and Analysis section.

\subsection{Decoding}

We use a transformer decoder that utilizes $H_{\text{fuse}}$ to generate text autoregressively. 
\begin{align}
    p(Y_t|Y_{<t}, X_{\text{lang}}, X_{\text{img}}, X_{\text{kg}}) = \text{Decoder}(Y_{<t}, H_{\text{fuse}})
\end{align}

\section{Experiments}
\subsection{Dataset}

We evaluate our method on the ScienceQA benchmark \citep{lu2022learn}.
It comprises of $21208$ multiple-choice questions with multimodal contexts, sourced from the science curriculum.
% ScienceQA stands out as the first large-scale multimodal QA dataset to include annotated letures and explanations for the answers. 
It covers substantial domain diversity, spanning $3$ subjects, $26$ topics, $127$ categories and $379$ skills.
ScienceQA provides us with an in-house training, dev and test split containing
$12726$, $4241$ and $4241$ samples respectively.
% We split the existing training set to $12726$ train and $4241$ dev samples.
% We evaluate the model on the available test set that contains $4241$ samples.

\subsection{Baseline Comparisons} 

We choose the following baselines, (i) VQA models \citep{kim2021vilt, lu2021iconqa, li-etal-2020-bert-vision}, (ii) Models with similar backbones \citep{khashabi-etal-2020-unifiedqa, lu2022learn,zhang2023multimodal, wang2023tsciq}, (iii) Parameter-efficient finetuned LLMs \citep{zhang2023llamaadapter, luo2023cheap}, and (iv) the GPT family and GPT-assisted models \citep{chatgpt, openai2023gpt4, liu2023visual}.

\begin{table*}[!t]
    \centering
    \fontsize{9}{10}\selectfont
    \begin{tabular}{l|r|cccccccc|c}
        \hline
        \textbf{Model} & \textbf{Size} & \textbf{NAT} & \textbf{SOC} & \textbf{LAN} & \textbf{TXT} & \textbf{IMG} & \textbf{NO} & \textbf{G1-6} & \textbf{G7-12} & \textbf{Avg} \\ \hline
        Human Average & - & 90.23 & 84.97 & 87.48 & 89.60 & 87.50 & 88.10 & 91.59 & 82.42 & 88.40 \\ \hline
        ViLT \citep{kim2021vilt} & 112M & 60.48 & 63.89 & 60.27 & 63.20 & 61.38 & 57.00 & 60.72 & 61.90 & 61.14 \\
        Patch-TRM \citep{lu2021iconqa} & 90M & 65.19 & 46.79 & 65.55 & 66.96 & 55.28 & 64.95 & 58.04 & 67.50 & 61.42 \\
        VisualBERT \citep{li-etal-2020-bert-vision} & 111M & 59.33 & 69.18 & 61.18 & 62.71 & 62.17 & 58.54 & 62.96 & 59.92 & 61.87 \\ \hline
        UnifiedQA\textsubscript{Base} \citep{khashabi-etal-2020-unifiedqa} & 223M & 68.16 & 69.18 & 74.91 & 63.78 & 61.38 & 77.84 & 72.98 & 65.00 & 70.12 \\
        UnifiedQA\textsubscript{Base} w/ CoT \citep{lu2022learn} & 223M & 71.00 & 76.04 & 78.91 & 66.42 & 66.53 & 81.81 & 77.06 & 68.82 & 74.11 \\
        MM-CoT\textsubscript{T5-Base} \citep{zhang2023multimodal} & 223M & 87.52 & 77.17 & 85.82 & 87.88 & 82.90 & 86.83 & 84.65 & 85.37 & 84.91 \\   
        MM-CoT\textsubscript{FLAN-T5-Base} & 250M & 91.5 & 74.92 & 90.09 &91.69 & 84.28 & 90.52 & 88.14 & 87.01 & 87.74 \\ 
        MM-T-SciQ\textsubscript{Base} \citep{wang2023tsciq} & 223M & 91.52 & 91.45 & 92.45 & 91.94 & 90.33 & 92.26 & 92.11 & 91.10 & 91.75 \\\hline
        
        % MM-CoT\textsubscript{FLAN-T5-Base} \citep{zhang2023multimodal} &  & 87.52 & 77.17 & 85.82 & 87.88 & 82.90 & 86.83 & 84.65 & 85.37 & 84.91 \\
        LLaMA-Adapter \citep{zhang2023llamaadapter} & 6B (1.2M) & 84.37 & 88.30 & 84.36 & 83.72 & 80.32 & 86.90 & 85.83 & 84.05 & 85.19\\ 
        LaVIN-13B \citep{luo2023cheap} & 13B (5.4M) & 89.88 & 94.49 & 89.92 & 88.95 & 87.61 & 91.85 & 91.45 & 89.72 & 90.83 \\ \hline

        GPT-3.5 w/ CoT \citep{chatgpt} & \textgreater175B & 75.44 & 70.97 & 78.09 & 74.68 & 67.43 & 79.93 & 78.23 & 69.68 & 75.15 \\
        GPT-4 w/ CoT \citep{openai2023gpt4} & \textgreater175B & 85.48 & 72.44 & 90.27 & 82.65 & 71.49 & 92.89 & 86.66 & 79.04 & 83.99 \\
        LLaVa (GPT-4) \citep{liu2023visual} & 13B & 91.56 & \textbf{96.74} & 91.09 & 90.62 & 88.99 & 93.52 & 92.73 & 92.16 & 92.53 \\ \hline

        KAM-CoT\textsubscript{T5-Base} \textbf{(Ours)} & 223M & 93.21 & 92.21 & 90.64 & 93.21 & \textbf{93.26} & 91.50 & 92.51 & 92.42 & 92.48  \\
        KAM-CoT\textsubscript{FLAN-T5-Base} \textbf{(Ours)} & 250M & \textbf{94.76} & 92.24 & \textbf{93.36} & \textbf{94.53} & 93.16 & \textbf{94.15} & \textbf{94.24} & \textbf{93.21} & \textbf{93.87}  \\ \hline
    \end{tabular}
    \caption{Comparing the results against baselines. Here, Size = size of the backbone model, 
    NAT = Natural Science, SOC = Social Science, LAN = Language Science, TXT = Text
    context, IMG = Image context, NO = No context, G1-6 = Grade 1 to 6, G7-12 = 
    from Grade 7 to 12. Segment 1 compares against 
    the human average. Segment 2 shows the performance of chosen VQA baselines.
    Segment 3 has models whose backbone sizes are comparable to ours.
    In Segment 4, we show parameter-efficient finetuned versions of
    larger models, and the number of trainable parameters are 
    provided inside parantheses. Segment 5 has the performance of the GPT family. 
    MM-CoT\textsubscript{FLAN-T5-Base} here has been given caption as context along with the vision features.
    Results, other than ours and MM-CoT\textsubscript{FLAN-T5-Base}, are taken from respective papers 
    and the ScienceQA leaderboard.}
    \label{table:results}
\end{table*}

\subsection{Training Details}
The size of the proposed model is $254$M with T5-Base and $280$M with 
FLAN-T5-Base.
All our experiments are run on a single NVIDIA A100 40G GPU.
We train our models for $20$ epochs, and also evaluate them after each,
with ScienceQA's dev split. We use a learning rate of 5e-5 and batch-size of 1, a maximum input length of 512 tokens, and maximum output length of 512 and 64 tokens for rationale and answer generation respectively.
% The exact system details and arguments used for training and evaluation are presented in 
% Appendix \ref{appendix:1}.

\subsection{Experimental Setup} 
% To start with, we establish the effectiveness of the proposed fusion technique vs non-graph baselines. 
% This allows us to quantify the impact of introducing the graph-based information on the model's performance.
% In the graph-enabled setting, we enrich the model's understanding with contextual relationships and dependencies derived from the  external knowledge.
% By feeding the model with relevant graph features, we explore the potential synergy between graph and other modalities. 
% We also explore few other variations of fusing modalities.
% Details are provided in the Discussion section.
For our experiments, we discuss the effect of using different image encoders.
(i) CLIP \citep{DBLP:conf/icml/RadfordKHRGASAM21} aligns images and text into a common embedding space.
(ii) DETR \citep{10.1007/978-3-030-58452-8_13} leverages transformers to perform object detection and localization.
The chosen variants of DETR\footnote{\url{https://huggingface.co/facebook/detr-resnet-101-dc5}} and CLIP\footnote{\url{https://huggingface.co/google/vit-base-patch16-384}} are used without their classification heads, to provide patch embeddings of shape $(100, 256)$ and $(49, 2048)$, respectively.

We experiment with caption features as well, where captions are generated using ViT-GPT2.\footnote{\url{https://huggingface.co/nlpconnect/vit-gpt2-image-captioning}}
% We therefore, also evaluate the setting with context concatenated with the caption.
Yet another set of experiments use these captions for extracting graph nodes.
In this case, right after generating the possible entailments of the sample, we put the caption seperated by a white-space.
The grounding process then continues as discussed in the Method section.
We also experiment with both the above mentioned settings.

To encode the knowledge-graph we use two layers: a Relational Graph Attention layer 
\citep{busbridge2019relational}, followed by a Graph Convolutional layer 
\citep{DBLP:conf/iclr/KipfW17}, both implemented in PyTorch Geometric \citep{fey2019fast}. 
We refrain from using more than two graph layers as that 
might lead to a node forgetting its own identity \citep{Li_Han_Wu_2018}.
The first graph layer uses $768$ input and output features,
matching the language encoder's
embedding dimension size. 
It is also provided with the number of possible relations, $34$
and the edge embedding size, $64$.
Next, the Graph Convolution layer is given only the input and output
feature sizes, both being set at $768$.
As mentioned in the Method section, for representing the edges, we learn an 
embedding table in the training process.
Given an integer for the edge-type, it produces an embedding, $e_{edge} \in \mathbb{R}^{64} $ 
for that edge, and is fed to the graph-encoder.

Our approach uses T5-Base \citep{raffel2020exploring}
as its backbone. The well defined encoder-decoder architecture
gives a good entry-point to introduce other modalities.
To ensure the applicability of our approach to other language models,
we conduct experiments and present results on the instruction-tuned FLAN-T5-Base 
\citep{chung2022scaling} also.

\subsection{Results}
To assess the effectiveness of our model, we use two evaluation metrics: average accuracy and RougeL \citep{lin-2004-rouge}.
Average accuracy quantifies the model's correctness in predicting the correct answer, and is treated as the primary metric for evaluating the quality of our method.
We use the RougeL metric to to compare the generated rationale to the human reference,
as done in \citet{zhang2023multimodal}.
ScienceQA contains multiple groups, that enables us to compare group-wise accuracies, giving an insight to the model's strengths and limitations within each group, which is valuable in understanding how the model generalizes across content areas.

For a fair and consistent evaluation, we obtain the scores of the baseline models directly from their respective research papers.
Additionally, we take scores from the ScienceQA leaderboard\footnote{\url{https://scienceqa.github.io/leaderboard.html}} for closed-source models. This enables us to make informed assessments of our model's contributions in comparison to existing state-of-the-art. Table \ref{table:results} shows the main results. 
Our model outperforms all other known approaches under $300$M and does not use any very large auxiliary model.
With FLAN-T5-Base as the backbone, we achieve a RougeL score of $98.40$ and an average accuracy of $93.87$, which is well above the performance of GPT3.5 ($75.17\%$), and also surpasses LLaVa ($92.53\%$) by $1.34\%$.
This conceretely establishes that our proposed method is superior compared to other approaches including LLMs, while being under $300$M parameters.
% Unfortunately, we did not have the compute to perform experiments with the larger variants of our chosen backbones.

A closer look into Table \ref{table:results} reveals that questions about Natural Science, Social Science and Language Science see a boost compared to the baselines.
The same is also observed for No-Context questions.
ConceptNet is expected to aid with these kind of questions, which is visible here clearly.
% We examine a few selected samples in Appendix \ref{appendix:3}.

We conduct further experiments and ablation studies to delve deeper into the performance and robustness of our proposed model.
We also explore the effects of varying the individual modalities and encoders.
We explore more fusion methods in the Additional Fusion Mechanisms subsection.

Unless explicitly mentioned, all experiments are trained and evaluated for 20 epochs,
and then tested on the test-split.

\begin{table}[]
    \centering
    \begin{tabular}{c|c|c|c}
    \hline
    \textbf{Image Features} & \textbf{Feature Size} & \textbf{RougeL} & \textbf{Avg. Acc} \\ \hline
    DETR & (100, 256) & \textbf{98.29} & \textbf{91.65} \\
    CLIP & (49, 2048) & 98.15 & 91.02 \\ \hline
    \end{tabular}
    \caption{Comparative results using different image encoders with T5-Base. DETR outperforms the CLIP based encoding.}
    %\caption{Image. Both settings were trained and evaluated for 20 epochs, then tested on the test split.}
    \label{table:image-features}
\end{table}
Table \ref{table:image-features} shows the effect of 
using different image encoders. DETR 
gives a marginal improvement $(0.63\%)$ over CLIP features, despite
having a smaller feature size ($74$k floats lesser) per sample,
making it our default choice.

\begin{table}[]
    \centering
    \begin{tabular}{c|c|c}
    \hline
    \textbf{Method} & \textbf{RougeL} & \textbf{Avg. Acc} \\ \hline
    without captions & 98.29 & 91.65 \\
    with captions as context & 98.33 & 92.45 \\
    captions for node extraction & 98.31 & 91.84 \\
    \textbf{captions for nodes + context} & \textbf{98.32} & \textbf{92.48} \\ \hline
    \end{tabular}
    \caption{Summary of results that showcase different approaches using captions with T5-Base.}
%    \caption{Test results with different setting involving the image captions. All settings use T5-Base, DETR image embeddings, trained and evaluated for 20 epochs, and tested with the test split.}
    \label{table:captions}
\end{table}
We observe from Table \ref{table:captions}, captions concatenated 
with the context gave a boost to both the rationale and the accuracy scores. 
In another setting where captions are concatenated with the context, and then used
to extract nodes, shows a marginal boost over not using them at all 
$(91.65 \rightarrow 91.84)$, but also with a very little fall in the 
RougeL score $(0.02)$.

The final combination, where captions are added to the context and also used for extracting 
node embeddings, turns out to be the best setting for average accuracy. 

\begin{table}[]
    \centering
    \begin{tabular}{c|c|c}
    \hline
    \textbf{Number of nodes} & \textbf{RougeL} & \textbf{Avg. Acc} \\ \hline
    50 & 97.78 & 88.66 \\
    100 & 97.84 & 88.85 \\
    200 & \textbf{97.85} & \textbf{89.51} \\ \hline
    \end{tabular}
    \caption{Effect of varying the number of nodes in a graph with T5-Base as the backbone.}
    %\caption{Effect of varying the number of nodes in a graph. All three experiments were performed using T5-Base, DETR embeddings, trained and evaluated for 10 epochs, then tested on the test split.}
    \label{table:nodes}
\end{table}
We study the effect of taking the top 50, 100 and 200 nodes. 
If the node extraction process yields a smaller number of nodes,
they are zero-padded to the minimum number.
To expedite these experiments with varying number of nodes, and to reduce GPU
consumption, we limit training to 10 epochs.
Limiting the maximum number of nodes has a proportional effect on the accuracy.
Table \ref{table:nodes} shows the trend that more nodes help the model 
reason and choose better. Although we could not perform exhaustive experiments 
with higher number of nodes, we anticipate that the performance would
saturate and might even decline beyond a certain 
threshold. We defer this aspect to future research.

\begin{table}[]
    \centering
    \begin{tabular}{c|c|c|c|c}
    \hline
    \textbf{Image} & \textbf{Captions} & \textbf{KG} & \textbf{RougeL} & \textbf{Avg. Acc} \\ \hline
    \ding{55} & \ding{55} & \ding{55} & - & 83.42\tablefootnote{Results are taken from \cite{zhang2023multimodal}\label{ablation: res}} \\
    \ding{51} & \ding{55} & \ding{55} & - & 85.85 \textsuperscript{\ref{ablation: res}}\\
    \ding{51} & context & \ding{55} & 97.27 & 87.74 \\
    \ding{51} & context & \ding{51} & 98.34 & 92.62 \\
    \textbf{\ding{51}} & \textbf{context, nodes} & \textbf{\ding{51}} & \textbf{98.40} & \textbf{93.87} \\ \hline
    \end{tabular}
    \caption{Ablation study on the KAM-CoT framework, using FLAN-T5-Base.}
    %\caption{Ablation study with FLAN-T5-Base as the backbone and DETR embeddings, trained and evaluated for 20 epochs, and then tested. Results with * are taken from \cite{zhang2023multimodal}}
    \label{table:flan-t5}
\end{table}
Having explored the effects of various settings over the modalities, 
we perform ablation studies, with FLAN-T5-Base as the backbone.
The complete model amounts to a total of $279$M trainable parameters with the 
graph encoder included.
From Table \ref{table:flan-t5}, it is easily seen that just plugging in the graph
encoder gives an accuracy boost of $4.88\%$, totaling to $92.62$, which surpasses
the performance of LLaVA (Table \ref{table:results}) with $13$B parameters, and is
yet not the highest score we could reach.

As reported in the beginning of this section, the best out of all our 
experiments come with the captions as context + node extraction setting. 
With $280$M paramters, our achitecture has a RougeL score of $98.40$ 
and an average accuracy of $93.87$, with a model \textbf{47 times smaller} 
than its next best performer.

\section{Discussion and Analysis}
In this section, we examine, a few alternative fusion mechanisms, model convergence, and results using subset of train data.

\subsection{Additional Fusion Mechanisms}
\label{sec:other_fusion_methods}

\begin{table}[]
    \centering
    \begin{tabular}{c|c|c|c} \hline
    \textbf{Fusion Method} & \textbf{\# Parameters} & \textbf{RougeL} & \textbf{Avg. Acc} \\ \hline
    Fusion-1 & 254M & \textbf{98.29} & \textbf{91.65} \\ 
    Fusion-2 & 251M & 98.23 & 91.23 \\ 
    Fusion-3 & 250M & 98.14 & 90.14 \\ \hline
    \end{tabular}
    \caption{Comparative performance of the varying fusion methods. Fusion-1 outperforms the other fusion methods.}
%    \caption{Results of the proposed fusion mechanisms. 
%    The number of parameters reported is with T5-Base as the backbone. Details about Fusion-2 and 3 are in the Discussion section.}
    \label{table:fusion}
\end{table}
Unlike the bottleneck-style~\cite{yasunaga2021qa} interaction between node embedding and other modalities, our fusion mechanisms have no such constraints.
Along with the proposed primary fusion method in the Fusion subsection, we experiment with two more settings.

%Exact mathematical formulation of both is given below.

\paragraph{2-step fusion (Fusion-2)} In the first stage, we fuse language-vision and language-KG features and get $H_{\text{img,kg}}$. 
Considering language as the primarily modality, we fuse it with $H_{\text{img,kg}}$ in the second stage.
\begin{equation}\begin{split}\label{eqn11}
    \lambda_{a} = \text{sigmoid}(H_{\text{img}}^{\text{attn}} W_{1} + H_{\text{kg}}^{\text{attn}} W_{2}) \in \mathbb{R}^{n \times d}
    \\
    H_{\text{img,kg}} = (1 - \lambda_{a}) \cdot H_{\text{img}}^{\text{attn}} + \lambda_{a} \cdot H_{\text{kg}}^{\text{attn}} \in \mathbb{R}^{n \times d}
\end{split}\end{equation}
\begin{equation}\begin{split}\label{eqn12}
    \lambda_{b} = \text{sigmoid}(H_{\text{img,kg}} W_{3} + H_{\text{lang}} W_{4}) \in \mathbb{R}^{n \times d}
    \\
    H_{\text{fuse}} = (1 - \lambda_{b}) \cdot H_{\text{img,kg}} + \lambda_{b} \cdot H_{\text{lang}} \in \mathbb{R}^{n \times d}
\end{split}
\end{equation}

\paragraph{1-step fusion (Fusion-3)}
In this approach we take the linear projection of $H_{\text{lang}}$, $H_{\text{img}}^{\text{attn}}$, $H_{\text{kg}}^{\text{attn}}$ and compute their weighted sum to merge all the modalities. 
\begin{equation}\begin{split}\label{eqn13}
    S_{\alpha} = H_{\text{lang}}W_{1}\;,\;
    S_{\beta} = H_{\text{img}}^{\text{attn}}W_{2}\;,\;
    S_{\gamma} = H_{\text{kg}}^{\text{attn}}W_{3}\;,\;
\end{split}\end{equation}
\begin{equation}\begin{split}\label{eqn14}
    \alpha_{ij}, \beta_{ij}, \gamma_{ij} = \text{softmax}([S_{\alpha_{ij}}, S_{\beta_{ij}}, S_{\gamma_{ij}}])
    \\    
    H_{\text{fuse}} = \alpha \cdot H_{\text{lang}} + \beta \cdot H_{\text{img}}^{\text{attn}} + \gamma \cdot H_{\text{kg}}^{\text{attn}} \in \mathbb{R}^{n \times d}
\end{split}\end{equation}
%In all these sets of equations, $W \in \mathbb{R}^{d \times d}$, $\alpha, \beta, \gamma \in [0, 1]^{n \times d}$ and sum to 1 element-wise.

\begin{figure}[]
    \begin{center}
        \includegraphics[width = 239pt]{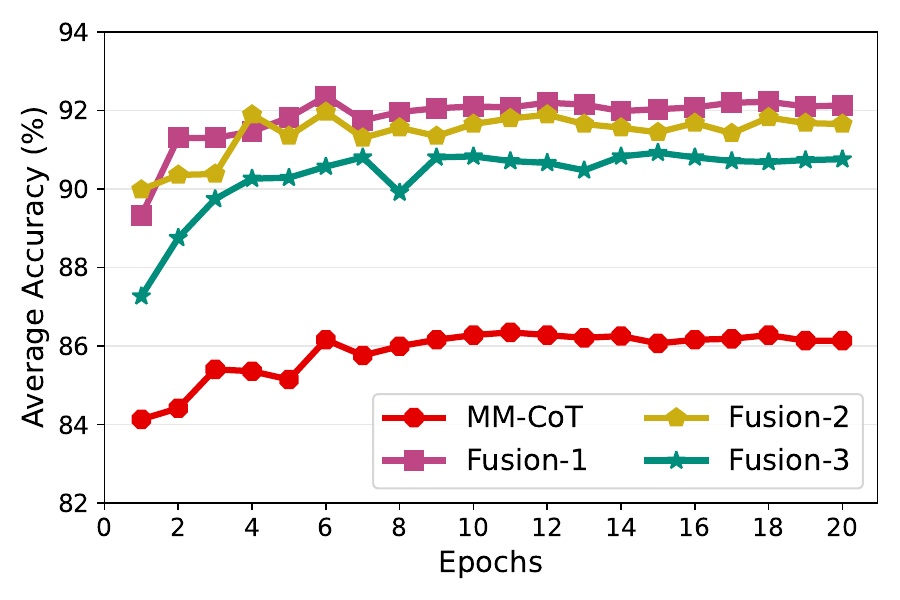}
    \end{center}
    \caption{Performance of the fusion mechanisms on the validation set, evaluated using T5-Base.}
    \label{figure:convergence}
\end{figure}

We summarise the results of these fusion mechanisms in Table~\ref{table:fusion} and find that Fusion-1 gives the best performance on ScienceQA test data.

\subsection{Comparing Model Convergence}
Figure \ref{figure:convergence} compares our model's convergence trend (with all fusion techniques) with 
MM-CoT~\cite{zhang2023multimodal} on the validation. We observe that the proposed method as well as MM-CoT converge at 10 epochs. Note that, the accuracy of the proposed approach starts much higher as compared to MM-CoT. Also, Fusion-1 demonstrates the highest accuracy, along with greater stability in comparison to others.

\subsection{Dataset Variation}
\begin{figure}[]
    \begin{center}
        \includegraphics[width = 239pt]{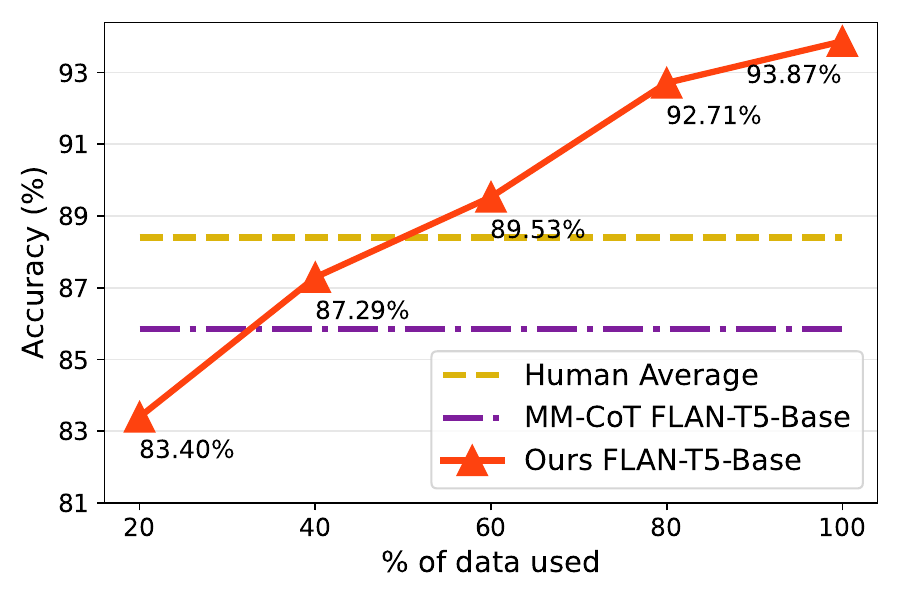}
    \end{center}
    \caption{Comparative performance using subsets of training data with MM-CoT\textsubscript{FLAN-T5-Base} (100\% training data, \citet{zhang2023multimodal}), and the human average.}
    \label{figure:fractions}
\end{figure}

To examine the scalability of the proposed model, we also train on 
subsets of the training data. These sets are made in the proportion of 20\%, 40\%, 60\% and 80\% of all the 
$12$k total training samples, preserving the distribution over the 26 topics. Figure~\ref{figure:fractions} shows that KAM-CoT surpasses human accuracy ($88.4\%$) even when trained with only $50\%$ of the training data. Surprisingly, the model outperforms the fully trained MM-CoT (Flan-T5\textsubscript{Base}) ($93.87\%$ vs $85.85\%$) with only $35\%$ of the training data.
The results highlight the model's generalization ability with little training data.

We also evaluate the model with A-OKVQA dataset. The proposed model outperforms the baseline by 3.67\%. 
% We refer the readers for detailed results in Appendix~\ref{app: appendix-2}.

\section{Conclusion}
In this paper, we propose KAM-CoT, 
Knowledge Augmented Multimodal Chain of Thought reasoning, 
to enhance the reasoning capability and quality of answers from language models. 
We propose a framework that uses CoT reasoning, leverages knowledge graphs 
and other modalities for a comprehensive understanding of multimodal tasks. 
We provide a few possible methods to fuse these modalities.
We find that the incorporation of KG in the two-stage 
training process helps reduce hallucinations. With only 280M parameters at a time, 
our approach yields a new state-of-the-art having an accuracy 
93.87\%, outperforming GPT-3.5 by 18\% by and GPT-4 by 10\%. In the future, we want to further integrate specific knowledge-intensive 
domains, and also explore efficient fusion mechanisms. 
We would also like to scale our solution to larger models like the LLaMA family.

\bibliography{aaai24, custom}

\clearpage

\appendix

\section{Appendix 1}
\subsection{Reproducibility of Results}
\label{appendix:1}

The section provides additional information to reproduce the experiments and results. We provide the
compute specifications, libraries and utilities used to train the models.

\subsubsection{Compute Infrastructure}
We execute the experiments using the following compute specifications.
\begin{itemize}
    \item NVIDIA A100 40 GB GPU $\times 1$
    \item 128 GB RAM
\end{itemize}
\begin{table}[b!]
    \centering
    \begin{tabular}{l|l}
    \hline
    \textbf{Package} & \textbf{Version} \\ \hline
    evaluate & 0.4.0 \\
    huggingface-hub & $\geq$ 0.4.0 \\
    nltk & 3.8.1 \\
    numpy & $\geq$ 1.23.2 \\
    pandas & 1.4.3 \\
    rouge & 1.0.1 \\
    rouge\_score & 0.1.2 \\
    sentence-transformers & 2.2.2 \\
    torch & 1.13.1+cu117 \\
    torch-cluster & 1.6.1+pt113cu117 \\
    torch-geometric & 2.3.1 \\
    torch-scatter & 2.1.1+pt113cu117 \\
    torch-sparse & 0.6.17+pt113cu117 \\
    torch-spline-conv & 1.2.2+pt113cu117 \\
    transformers & 4.21.1 \\ \hline
\end{tabular}
    \caption{Libraries and the corresponding versions.}
    \label{table:packages}
\end{table}
We use \textit{python 3.8.10} in our experiments. In Table \ref{table:packages}, we list the libraries along with the versions.

\subsubsection{Trainer Configuration}
To facilitate and make the training process seamless, we use the Seq2Seq HuggingFace trainer\footnote{\url{https://huggingface.co/docs/transformers/main_classes/trainer}}. We provide the training arguments in Table \ref{table:arguments}. We use 
a max\_length of 512 tokens for the input sequence, 512 and 64 output tokens for rationale and answer generation respectively.
\begin{table}[h!]
    \centering
    \begin{tabular}{l|c}
    \hline
    \textbf{Argument} & \textbf{Value} \\ \hline
    model & google/flan-t5-base \\
     & allenai/unifiedqa-t5-base \\
    per\_device\_train\_batch\_size & $1$ \\
    per\_device\_eval\_batch\_size & $1$ \\
    learning\_rate & $5 \times 10^{-5}$ \\
    eval\_accumulation\_steps & None \\
    weight\_decay & $0.01$ \\
    num\_train\_epochs & $20$ \\ \hline
    % save\_strategy & epochs \\ \hline
    \end{tabular}
    \caption{Seq2SeqTrainer Arguments.}
    \label{table:arguments}
\end{table}

\begin{figure}[t!]
    \begin{center}    
        \includegraphics[width = 239pt]{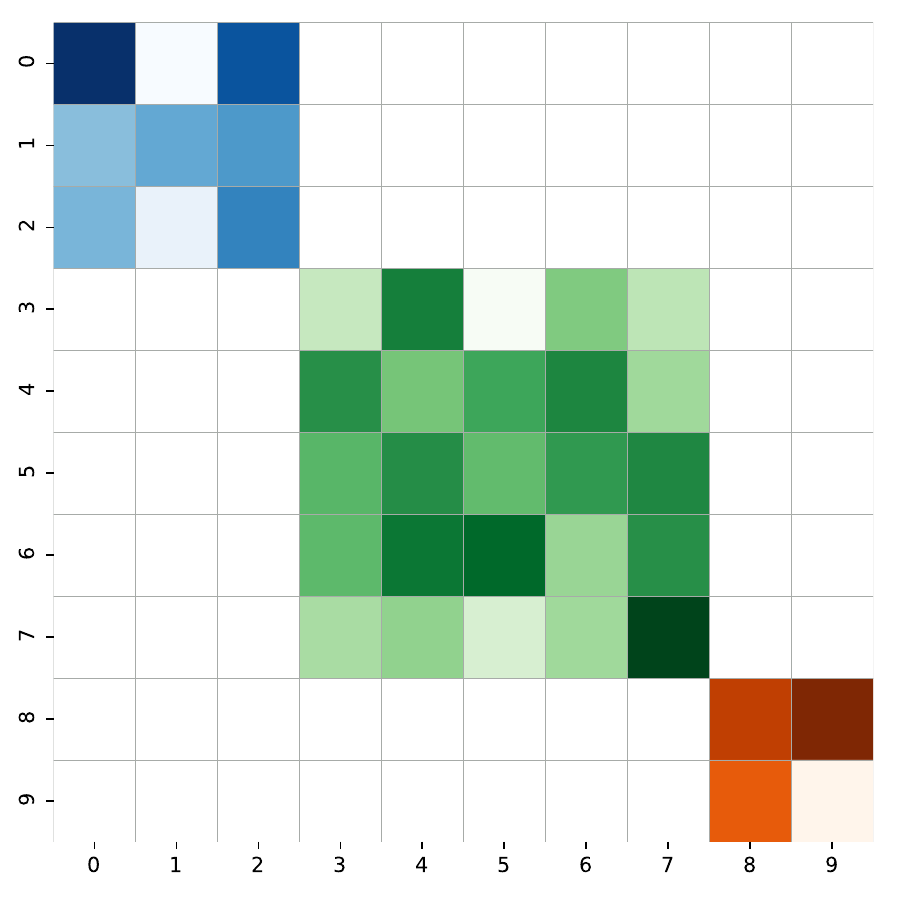}
    \end{center}
    \caption{A pictorial representation of batches with graph data. Here, the $3 \times 3$ blocks in the top-left, the $5 \times 5$
    blocks around the center and the lower-right $2 \times 2$ blocks are 3 
    disjoint graphs. The connections between nodes are indicated by the 
    intensity of fill in corresponding cell of the adjacency matrix. 
    Note that three nodes in the first graph do not have any relation 
    with any node beyond themselves. A similar observation can be made in the 
    other two graphs as well. Although disjoint, they can be put into a 
    single adjacency matrix as depicted above.}
    \label{figure:batching}
\end{figure}
\subsection{Batching with Graph Data}
When a set of disjoint graphs are included in a single batch, pyTorch geometric
collates them into a single graph. These disjoint graphs, represented as matrices, are stacked along the diagonal to make a `super' 
adjacency matrix. We show an example of stacking 3 graphs into a single one in Figure~\ref{figure:batching}.
Batching using PyTorch Geometric's dataloader is very convenient. However, this would lead to implement a custom data loader to be compatible
with the Seq2SeqTrainer. Instead, we designed a custom data collator, based on HuggingFace's 
DataCollatorForSeq2Seq. The custom collator merges graphs in the batch, along with batching the text data.

\subsection{Exploring a Few Generated Samples}
\label{appendix:3}
We examine some examples that are labelled incorrect by the baseline
\citep{zhang2023multimodal} method, but corrected by our model. Observe 
that the addition of the knowledge triples helps the model to generate 
effective rationales and answers.
\begin{figure}[t!]
    \begin{center}
        \includegraphics[width = 239pt]{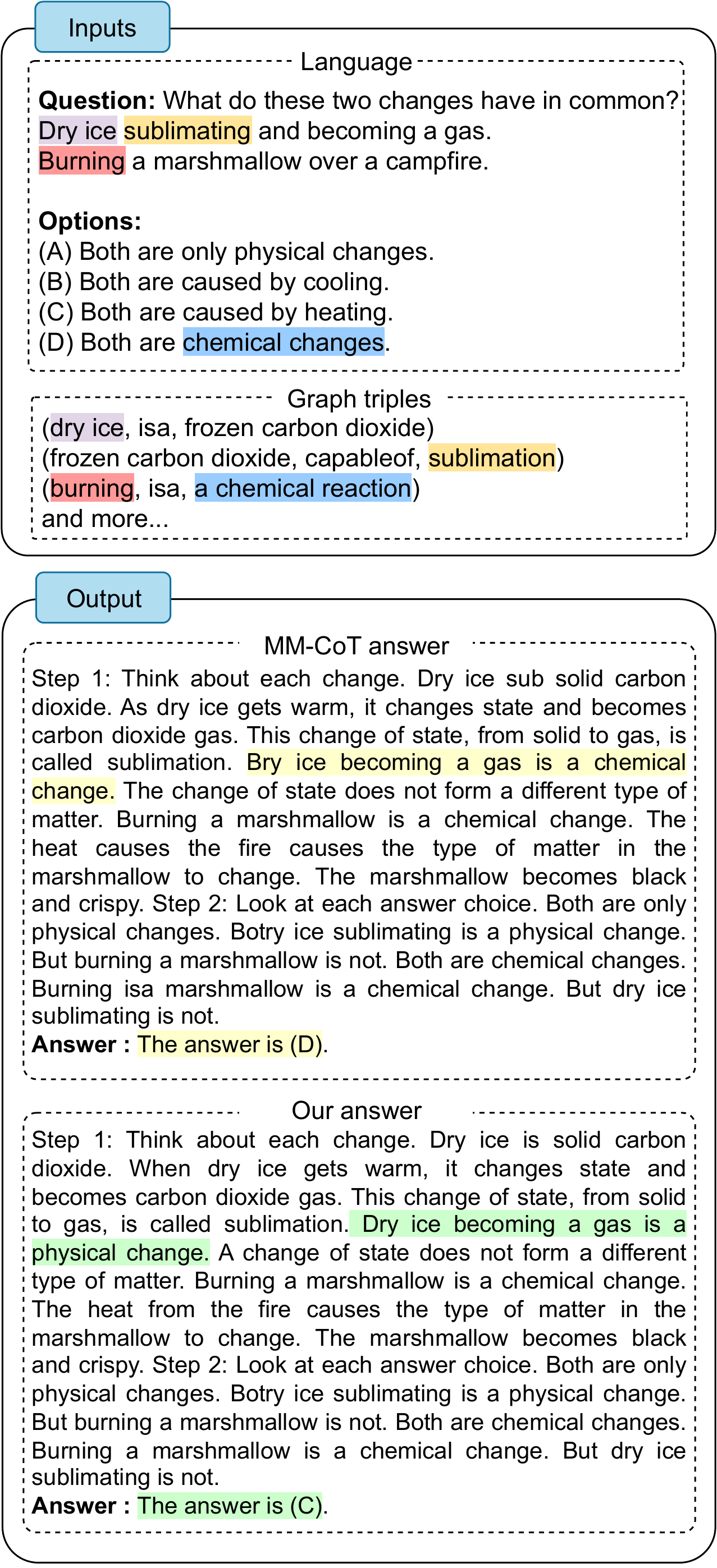}
    \end{center}
    \label{figure:sample1}
\end{figure}
In the following examples, the upper block shows the inputs provided to the models. The graph triples are the extracted relations based on the given input. 
%However, the graph triples are used by our model only, not the baseline.
The highlighted key-words in the text correspond to the grounded nodes from the ConceptNet. 

\begin{figure}[t!]
    \begin{center}
        \includegraphics[width = 239pt]{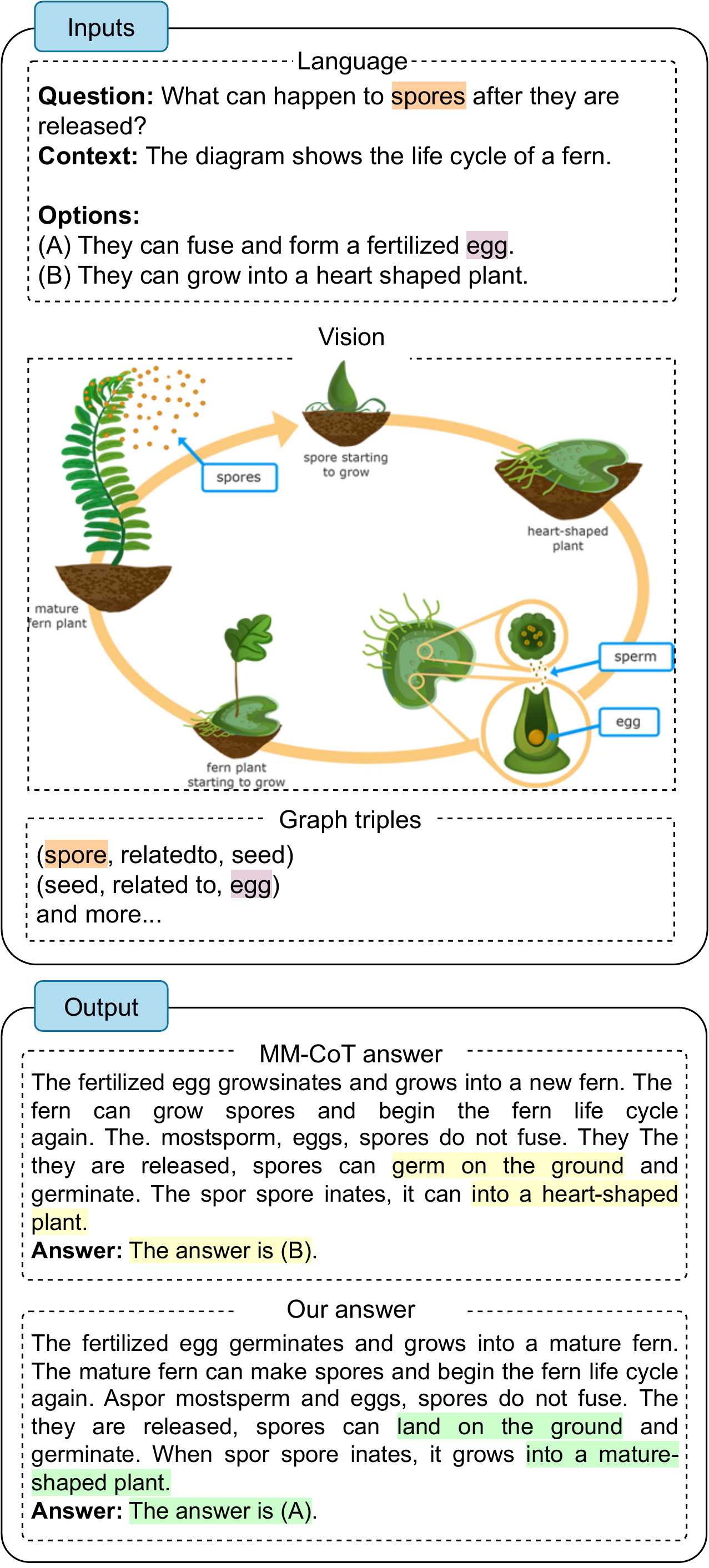}
    \end{center}
    \label{figure:sample2}
\end{figure}
\section{Appendix 2}
\begin{table}[h!]
    \centering
    \begin{tabular}{l|c|c}
    \hline
    \textbf{Method} & \textbf{Size} & \textbf{Val Acc} \\ \hline
    LXMERT \citep{tan-bansal-2019-lxmert} & 223M & 51.4 \\
    KRISP \citep{9577534} & 116M & 51.9 \\ \hline
    MM-CoT\textsubscript{FLAN-T5-Base} & 250M & 55.98 \\ 
    KAM-CoT\textsubscript{FLAN-T5-Base} \textbf{(Ours)} & 250M & 59.65 \\ 
    GPV-2 \citep{kamath2022webly} & 370M & 60.3 \\ \hline
    Prophet \citep{shao2023prompting} & 175B & 76.4 \\
    PromptCAP \citep{hu2022promptcap} & 175B & 73.2 \\ \hline
    \end{tabular}
    \caption{Validation set performance for A-OKVQA. Here, size refers to the size of the backbone model.
    Results for all methods, except ours and MM-CoT, are taken from respective papers.}
    \label{table:aokvqa}
\end{table}
\label{app: appendix-2}
\subsection{Performance on other datasets}
To validate the effectiveness of our proposed method, we also evaluate our
model against the A-OKVQA \citep{aokvqa} dataset. It is a large, external-knowledge based VQA dataset, which
contains 25k samples which were all made to ensure that they would need external knowledge for answering. 
The dataset consists of 17k training, 1k validation and 
6.7k test samples. 

To train our model, we made use of the best setting, 
with Flan-T5\textsubscript{Base}, DETR image embeddings and captions.
Our method is evaluated on the multiple choice task 
where each question has 4 choices. We compare the proposed method with models having less than 1B parameters only.
We also fine-tune MM-CoT (Flan-T5\textsubscript{Base}) \citep{zhang2023multimodal}, 
to establish as a baseline for this task.
We summarize the results in Table \ref{table:aokvqa}. Observe that KAM-CoT surpasses the baseline by $3.67\%$. 
However it is behind the best performer by a small margin of $0.65\%$. For completeness, we also provide the results of the models with 175B parameters.

\end{document}